\documentclass{article}

\usepackage{microtype}
\usepackage{graphicx}
\usepackage{subcaption}
\usepackage{booktabs} 

\usepackage{hyperref}

\usepackage{algorithm}
\usepackage{algpseudocode} 

\usepackage[preprint]{icml2026}

\usepackage{amsmath}
\usepackage{amssymb}
\usepackage{mathtools}
\usepackage{amsthm}
\usepackage{multirow}
\usepackage{makecell}

\usepackage[capitalize,noabbrev]{cleveref}

\theoremstyle{plain}

\theoremstyle{definition}

\theoremstyle{remark}

\usepackage[textsize=tiny]{todonotes}

\icmltitlerunning{Past- and Future-Informed KV Cache Policy with Salience Estimation in Autoregressive Video Diffusion}

\begin{document}

\twocolumn[
      \icmltitle{Past- and Future-Informed KV Cache Policy with Salience Estimation \\ in Autoregressive Video Diffusion}

      


  \icmlsetsymbol{equal}{*}

  \begin{icmlauthorlist}
    \icmlauthor{Hanmo Chen}{hyy}
    \icmlauthor{Chenghao Xu}{hh}
    \icmlauthor{Xu Yang}{xd}
    \icmlauthor{Xuan Chen}{xd}
    \icmlauthor{Cheng Deng}{xd}
  \end{icmlauthorlist}

  \icmlaffiliation{hyy}{Hangzhou Institute of Technology, Xidian University, Hangzhou, China}
  \icmlaffiliation{xd}{Xidian University, Xi'an, China}
  \icmlaffiliation{hh}{Hohai University, Nanjing, China}

  \icmlcorrespondingauthor{Xu Yang}{xuyang.xd@gmail.com}
  
  \icmlkeywords{Machine Learning, ICML}

    \resizebox{\linewidth}{!}{
    \includegraphics{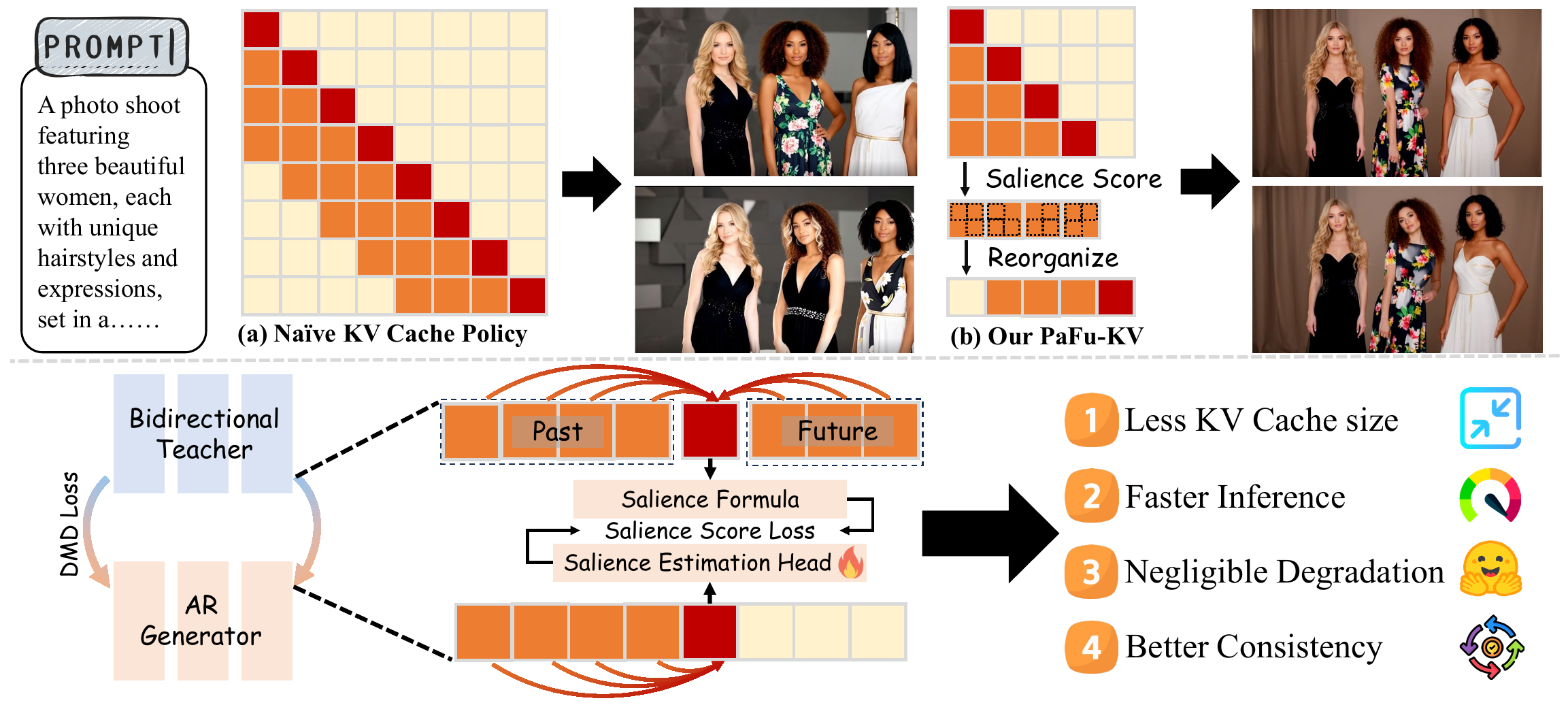}}
    \captionof{figure}{By distilling both past and future contextual information from a bidirectional teacher model, our PaFu-KV retaining KV Cache token with high salience score, achieving less KV Cache size and faster inference with negligible degradation on generation quality.}
    \label{fig:fig0}
    \vspace{0.4cm}

  \vskip 0.3in
]



\printAffiliationsAndNotice{}  

\begin{abstract}


Video generation is pivotal to digital media creation, and recent advances in autoregressive video generation have markedly enhanced the efficiency of real-time video synthesis. However, existing approaches generally rely on heuristic KV Cache policies, which ignore differences in token importance in long-term video generation. This leads to the loss of critical spatiotemporal information and the accumulation of redundant, invalid cache, thereby degrading video generation quality and efficiency. To address this limitation, we first observe that token contributions to video generation are highly time-heterogeneous and accordingly propose a novel Past- and Future-Informed KV Cache Policy (PaFu-KV). Specifically, PaFu-KV introduces a lightweight Salience Estimation Head distilled from a bidirectional teacher to estimate salience scores, allowing the KV cache to retain informative tokens while discarding less relevant ones. This policy yields a better quality-efficiency trade-off by shrinking KV cache capacity and reducing memory footprint at inference time. Extensive experiments on benchmarks demonstrate that our method preserves high-fidelity video generation quality while enables accelerated inference, thereby enabling more efficient long-horizon video generation. Our code will be released upon paper acceptance.

\end{abstract}

\section{Introduction}

Video generation plays a crucial role in digital media creation~\cite{xue2025human,ma2025controllable}, driven by recent Diffusion Transformer (DiT)-based models, such as Wan~\cite{wan2025wan}, Hunyuan~\cite{kong2024hunyuanvideo}, and CogVideo~\cite{yang2024cogvideox}, which have demonstrated remarkable progress in producing impressive videos~\cite{yang2025videograin}. However, their reliance on bidirectional attention incurs high memory overhead and long generation latency~\cite{liang2024looking}, which in turn degrades quality in long video generation and limits support for immersive, real-time interaction. To address this limitation, recent work~\cite{huang2025self, yang2025longlive,rajasegaran2025empirical} integrates the autoregressive (AR) paradigm with Diffusion Transformers (DiT)~\cite{peebles2023scalable} for sequential video generation using KV caching, enabling low-latency and real-time interactive video generation~\cite{yang2025longlive, shin2025motionstream}.

Traditional autoregressive~\cite{ren2025beyond,tian2024visual,yu2025videomar} video generation models adopt indiscriminate KV retention and eviction strategies, such as first-in-first-out (FIFO), ignoring the importance of different tokens in long-horizon video generation. These heuristic key-value (KV) caching strategies lead to the loss of critical spatiotemporal contextual information, exacerbated error accumulation, and ultimately a dual degradation in generation quality and efficiency. Prior works~\cite{xiang2025macro, yi2025deep} have explored complementary strategies to mitigate such limitations, including sparse KV Cache retrieval at the semantic or prompt level, yet these methods invariably introduce extra computational overhead during inference. More importantly, such strategies suffer from an inherent myopia: they assess token importance solely based on the KV Cache of existing local neighbors, without accounting for a token’s potential influence on future generations and long-horizon historical KV Cache states, causing the performance gap that is to be bridged in this paper.

Our work addresses an important yet underexplored challenge in AR video diffusion: causal frame generation relies exclusively on past tokens, overlooking future frame correlations. This asymmetric temporal dependency limits the model's ability to reason over long horizons and often degrades long-range identity consistency, particularly in scenarios involving object occlusion or reappearance. To tackle this, we propose a token salience scoring criterion grounded in both past and future relevance, enabling precise and reliable KV Cache management. Specifically, high-salience tokens are strongly backed by past or future spatiotemporal context and exhibit high self-consistency, making them robust for long-term cache retention. In contrast, low-salience tokens lack stable global anchoring, are prone to representational drift, and pose a high-risk source of error accumulation. Building on this criterion, we leverage a large-scale DiT-based teacher model to capture long-range dependencies that causal AR models cannot access, thanks to its inherent bidirectional self-attention mechanism that simultaneously models past and future token relationships. We further introduce a spatialtemporal-balanced salience scoring strategy that explicitly disentangles attention contributions from past, present, and future interactions, thus enabling more accurate estimation of the global utility of tokens in long video generation.


Based on the aforementioned salience scoring strategy, we propose Past- and Future-informed KV Cache (PaFu-KV), a novel caching policy for AR video generation that significantly reduces KV Cache footprint while preserving generation fidelity. Specifically, as causal AR video diffusion models inherently lack access to future information during inference, we integrate a lightweight learnable Salience Estimation Head (SEH) into the DiT backbone to predict token-level salience scores, where the training is seamlessly incorporated into the teacher model with the data-free Distribution Matching Distillation paradigm~\cite{yin2024one, yin2024improved}, thereby uncovering the historical tokens actually required for future generation under optimal information flow. During inference, PaFu-KV leverages SEH-predicted token salience scores to retain informative tokens in the KV Cache and evict those with marginal generative contributions. By selectively preserving salient tokens and discarding low-salience ones, the model learns to retain key information for fidelity preservation while discarding redundant tokens to mitigate error amplification, yielding more stable and coherent long-horizon video generation. Notably, based on the key observation that salient tokens exhibit heavy overlap across layers, we can efficiently capture global salience information by introducing a lightweight salience score prediction only at a single specific layer, while preserving the model’s inference efficiency. Extensive experiments on public benchmark datasets validate the efficacy of our method, and our key contributions are summarized as follows:

\begin{itemize}
	\item We propose PaFu-KV, a novel past- and future-informed KV Cache policy that explicitly targets KV Cache size reduction for efficient AR video generation.
	\item We design a lightweight Salience Estimation Head (SEH) and train it via Distribution Matching Distillation in a data-free paradigm, distilling omnidirectional knowledge from a bidirectional teacher to accurately estimate token importance under causal generation constraints with minimal consumption.
    \item Extensive experiments demonstrate that PaFu-KV significantly reduces KV Cache size and accelerates inference while preserving high-quality video generation.
\end{itemize}

\section{Related Work}

\textbf{Autoregressive Video Diffusion.} To enable efficient long video generation~\cite{zhang2025packing, jiang2024videobooth,li2025stable}, recent studies have increasingly focused on AR video diffusion. In the early stage,~\citet{gao2024ca2} adopts a teacher-forcing scheme and proposed Ca2-VDM, while~\citet{chen2024diffusion} introduced Diffusion Forcing. Building on this paradigm, CausVid~\cite{yin2025slow} leverages DMD~\cite{yin2024one} to distill a few-step AR video generator from a pretrained bidirectional DiT-based teacher model with Diffusion Forcing~\cite{chen2024diffusion} framework. However, these approaches suffer from training-inference discrepancy, which leads to error accumulation during inference. Self Forcing~\cite{huang2025self} mitigates this issue by conditioning each frame on previously generated context, but still exhibits error accumulation and semantic drift when generating long videos. Drawing inspiration from StreamingLLM~\cite{xiao2023efficient} in the Large Language Model (LLM) domain,~\citet{yang2025longlive} proposed LongLive, which introduces a frame sink mechanism and extends training of AR video diffusion to substantially longer video sequences. Additionally, Reward Forcing proposed by~\citet{lu2025reward} augmented DMD with a reward function and applies frame sink with an Exponential Moving Average.~\citet{yu2025videossm} incorporated a state-space model~\cite{gu2024mamba} into AR video diffusion to jointly process global and local memory. Although effective at mitigating error accumulation and training-inference discrepancy, these methods largely overlook KV Cache optimization, which is essential for improving efficiency and reducing memory footprint in AR video diffusion.


\noindent\textbf{KV Cache in Large Language Models.}
Recent works in the LLM domain have explored improving the efficiency of AR generation by reducing the memory footprint and computational cost of the KV Cache. Early approaches employ attention-driven token selection to identify and retain important tokens, such as H2O~\cite{zhang2023h2o}, which preserves tokens with high cumulative attention scores, and SnapKV~\cite{li2024snapkv}, which exploits temporal locality through observation windows. Beyond token-level, hierarchical compression strategies have been proposed to better preserve long-range context. PyramidalKV~\cite{cai2024pyramidkv} progressively downsamples older KV entries into coarser representations, while CAKE~\cite{qin2025cake} introduces adaptive cascading eviction policies that account for layer-wise sensitivity. In contrast to layer-wise methods, Ada-KV~\cite{feng2024ada} explores head-wise KV Cache eviction by establishing a theoretical loss bound and allocating adaptive budgets across attention heads. In parallel, RazorAttention~\cite{tang2024razorattention} leverages retrieval heads to guide KV Cache compression by selectively preserving KV pairs likely to be accessed during decoding. Collectively, these methods aim to improve inference efficiency by approximating full attention through principled KV Cache compression, without modifying the underlying attention operator. While these methods are effective in LLM, the multi-step denoising nature of diffusion prevents their direct adoption. Thus, we focus on designing a KV Cache policy specifically tailored for AR video diffusion in this work.

\begin{figure*}[ht]
  \centering
  \includegraphics[width=\linewidth]{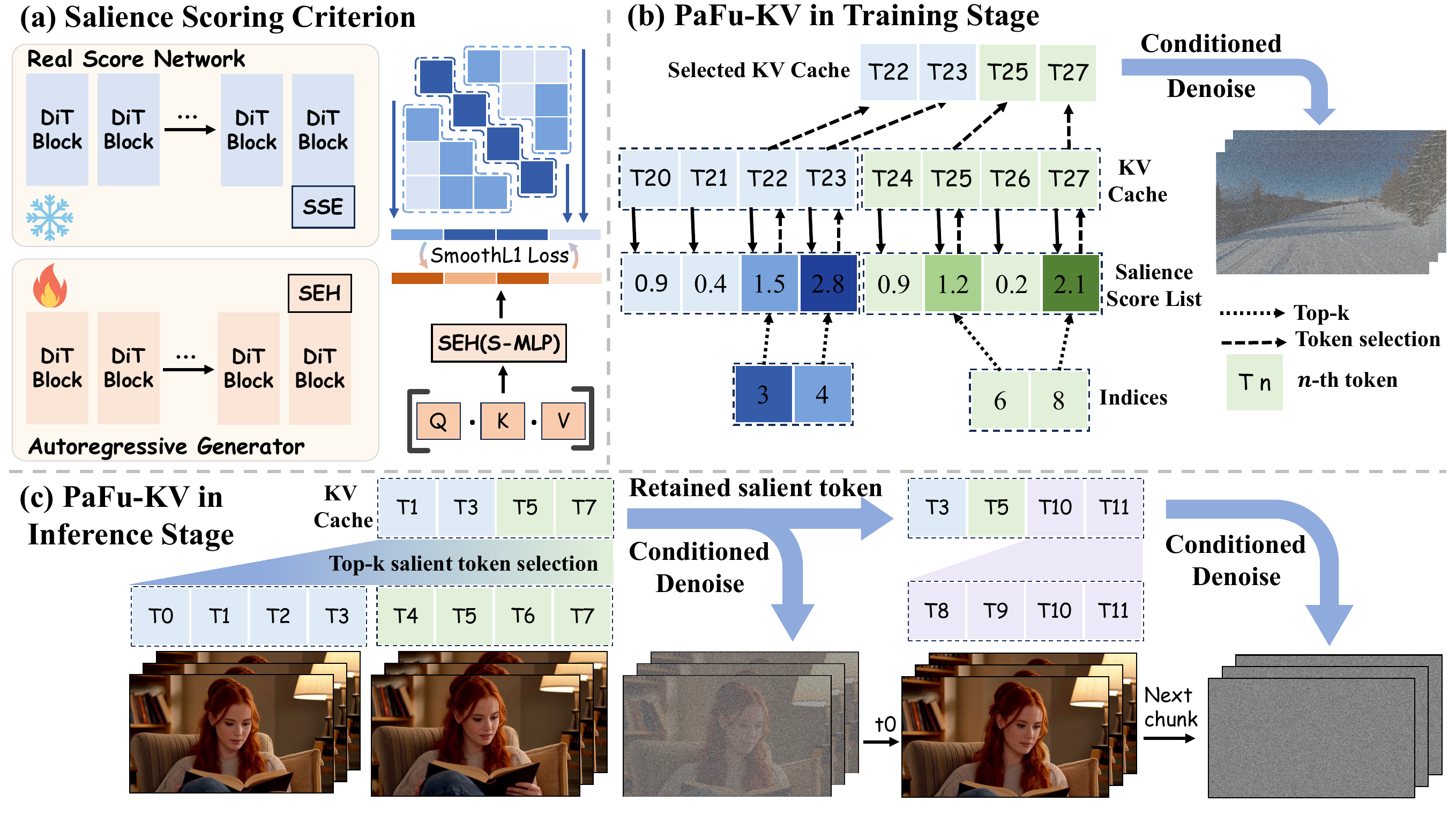}
  \caption{Overall framework of PaFu-KV. (a) Overview of the salience score criterion used during training. (b) Detailed training procedure of PaFu-KV. The salience score list is maintained in sync with the KV Cache. (c) Inference pipeline of PaFu-KV, where the KV Cache is maintained at a compact size via salience-based eviction.}
\label{fig:framework}
\end{figure*}

\section{Method}

\subsection{Preliminary}

\textbf{Autoregressive Video Diffusion.} In AR video diffusion models, the joint distribution over an $N$-frame video $x_{1:N}$ is factorized as
\begin{equation}
    p(x^{1:N})=\prod_{i=1}^Np(x^i\mid x^{<i}).
\end{equation}

A key challenge in AR video diffusion is the training-inference discrepancy. Self-Forcing mitigates this issue by introducing a self-rollout mechanism, which aligns training with inference via a diffusion-based conditional formulation. Specifically, each frame is generated by denoising Gaussian noise with a few-step diffusion model $G_\theta$, conditioned on previously generated frames via sliding-window attention with a fixed window size $L$ over timesteps $\{t_0, t_1, \ldots, t_T\}$. A KV Cache stores representations of recent frames and is maintained with a fixed capacity using a FIFO eviction policy. At each timestep $t_{j}$, the denoised $i$-th frame is re-perturbed via a forward diffusion process $\Psi$, yielding the input $x_{t_{j-1}}^i$ for the subsequent denoising step. This process can be formulated as:
\begin{equation}
\label{eq2}
    x_{t_{j-1}}^i=\Psi\left(G_\theta(x_{t_j}^i,t_j,KV),t_{j-1}\right),
\end{equation}
where $KV$ is the KV Cache from previous frames.

\textbf{Distribution Matching Distillation.} 
Distribution Matching Distillation (DMD) formulates few-step diffusion distillation as a distribution matching problem based on the reverse Kullback–Leibler (KL) divergence. It employs a few-step generator $G_\theta$, together with a frozen real score network and a continuously trained fake score network. Rather than explicitly computing the KL divergence, DMD optimizes the generator by directly using the gradient of the reverse KL objective with respect to $\theta$. By defining $x = G_\theta(z)$ with $z \sim \mathcal{N}(0, \mathbf{I})$, the gradient of the reverse KL divergence with respect to $\theta$ can be written as:
\begin{equation}
\nabla_{\theta} D_{\text{KL}} \!\!
=\!\!
\mathbb{E}_{z \sim \mathcal{N}(0, \mathbf{I})} \!
\!\left[\!
-\!\Big(
s_{\text{real}}(x)\! - \!s_{\text{fake}}(x)
\!\Big)
\frac{\partial G_\theta(z)}{\partial \theta}
\!\right],
\end{equation}
where $s_\mathrm{*}(x)=\nabla_x \log p_\mathrm{*}(x)$ denotes the score function. This formulation reveals that the optimization of $G_\theta$ is governed by the discrepancy between the real and fake score, thereby providing a principled direction toward the target data distribution.

\subsection{Spatialtemporal-balanced Salience Estimation}
\label{sec3.2}

As illustrated in Fig.~\ref{fig:framework}(a), our salience scoring criterion operates through two complementary components: (i) a Spatialtemporal-balanced Salience Estimation (SSE), detailed in this section, and (ii) the Salience Estimation Head (SEH), described in Sec.~\ref{sec3.3}. Prior studies~\cite{zhang2023h2o, singhania2024loki} demonstrate that attention weights effectively capture token importance in self-attention. Given input of $Q, K \in \mathbb{R}^{B \times N \times L \times D}$, where $B$, $N$, $L$, and $D$ denote the batch size, number of attention heads, sequence length and head dimension, respectively, attention weights are computed by $P=\mathrm{softmax}(QK^\top/\sqrt{D})$, where $P\in \mathbb{R}^{B \times N \times L \times L}$. We estimate key importance by taking the maximum attention weights over the query dimension, capturing the strongest interaction between each key and any query, where higher scores indicate stronger relevance. Consistent with prior works, removing tokens with low scores incurs negligible impact on generation quality. Furthermore, we observe a pronounced diagonal bias in $P$. Specifically, we collect key indices corresponding to maximal query responses and construct a $7 \times 7$ histogram by uniformly partitioning the index and token ranges, as shown in Fig.~\ref{fig1}(a). Our empirical analysis indicates that self-attention predominantly focuses on intra-frame or temporally local information. As a result, naïve global max-based aggregation over-selects local tokens while under-representing cross-frame interactions that are essential for temporal coherence in video generation. 

Consequently, our SSE is explicitly designed to mitigate this bias and better capture long-range temporal dependencies. We set the length of each block as $L_B$, and further decompose $P$ into three components based on each block: the upper triangular, the diagonal, and the lower triangular. For a query and key at position $i$ and $j$ respectively, the lower and upper boundaries for each block are defined as $\ell(i)=b(i)\cdot {L_B}$ and $h(i)=\ell(i)+{L_B}$, where $b(i)=\lfloor\frac{i}{{L_B}}\rfloor$ and $\lfloor \cdot \rfloor$ denotes the floor operation. Accordingly, we define the index sets corresponding to each component as
$\mathcal{I}_{\mathrm{low}}(j)=\{\, i \mid j < \ell(i) \,\}$,
$\mathcal{I}_{\mathrm{diag}}(j)=\{\, i \mid \ell(i) \leq j < h(i) \,\}$, and
$\mathcal{I}_{\mathrm{up}}(j)=\{\, i \mid j \geq h(i) \,\}$, respectively. We then take the maximum over the query dimension for each component and average across attention heads. The value for each components are computed as follows:
\begin{equation}
\label{eq4}
\begin{aligned}
\mathrm{low}_{b,j} &= \frac{1}{N} \sum_{n}
\left( \max_{i \in \mathcal{I}_{\mathrm{low}}(j)} P_{b,h,i,j} \right), \\
\mathrm{diag}_{b,j} &= \frac{1}{N} \sum_{n}
\left( \max_{i \in \mathcal{I}_{\mathrm{diag}}(j)} P_{b,h,i,j} \right), \\
\mathrm{up}_{b,j} &= \frac{1}{N} \sum_{n}
\left( \max_{i \in \mathcal{I}_{\mathrm{up}}(j)} P_{b,h,i,j} \right).
\end{aligned}
\end{equation}

Overall, we aggregate these values to define the salience score $s$, which captures a token’s strongest influence across past, present, and future interactions, as formulated below:
\begin{equation}
\label{eq5}
{s}_{b,j}=
\begin{cases}
\frac{\mathrm{diag}_{b,j}+\mathrm{low}_{b,j}}{2}, & 0\leq j<{L_B}, \\
\frac{\mathrm{up}_{b,j}+\mathrm{diag}_{b,j}+\mathrm{low}_{b,j}}{3}, & {L_B}\leq j<L-{L_B}, \\
\frac{\mathrm{diag}_{b,j}+\mathrm{up}_{b,j}}{2}, & L-{L_B}\leq j<L.
\end{cases}
\end{equation}
\subsection{Salience Estimation Head}
\label{sec3.3}

A fundamental challenge in the AR video diffusion is that future information is unobservable, while the salience score depends on both historical context and anticipated future utility. To this end, we introduce a learnable Salience Estimation Head (SEH) that infers salience score from existing representations while implicitly modeling future relevance.

\begin{figure}
  \centering
  \includegraphics[width=\linewidth]{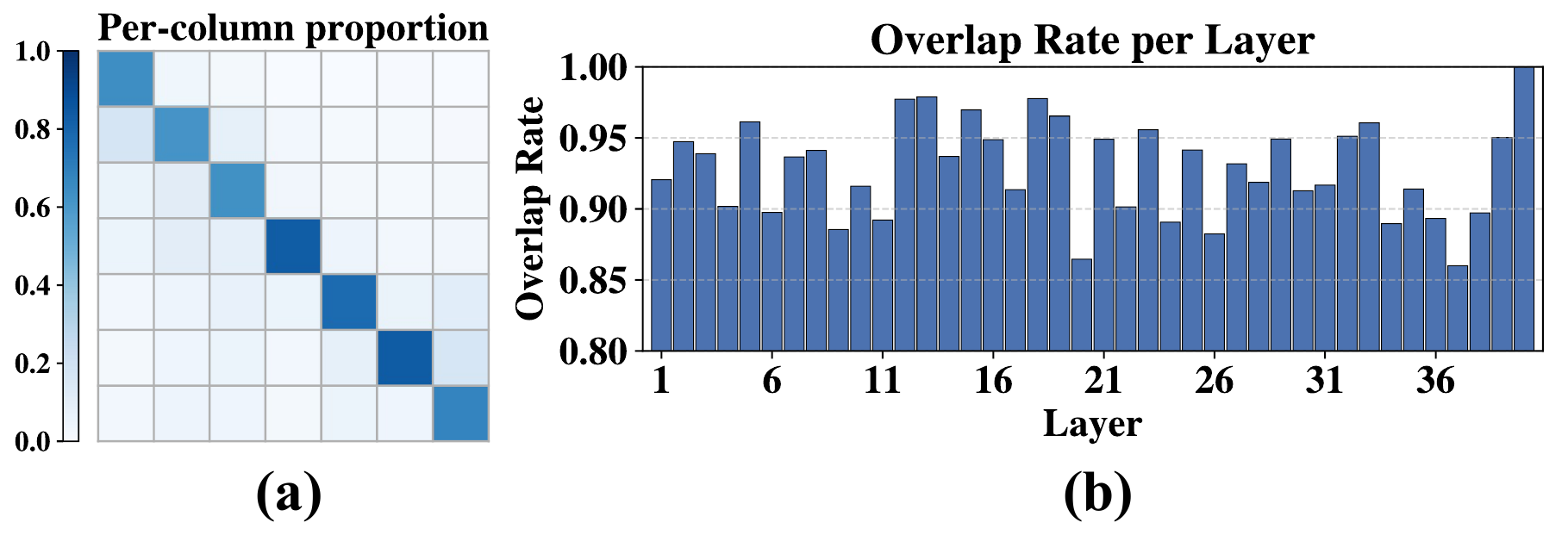}
  \caption{(a) A $7 \times 7$ count matrix constructed by uniformly partitioning key indices with maximal query responses and the token index range. (b) Overlap ratio of top-$k$ salient token indices between each intermediate Transformer layer and the final layer of Wan2.1-14B at the final denoising step $t_0$.}
\label{fig1}
\end{figure}

To validate this design and analyze the effective placement of salience estimation, we compute salience scores at each Transformer layer of Wan2.1-14B and extract the top-$k$ salient tokens, with $k = L/2$. Taking the final layer as a reference, we measure the overlap between its top-$k$ tokens and those from earlier layers. Since AR video diffusion retains the KV Cache only at the final denoising step $t_0$~\cite{huang2025self}, our analysis focuses on salience scores evaluated at $t_0$. As shown in Fig.~\ref{fig1}(b), the top-$k$ tokens across layers exhibit substantial overlap at $t_0$, indicating strong cross-layer consensus on token salience. Since the final layer operates on the most semantically stabilized representations, its top-$k$ token selection reliably approximates that of earlier layers with negligible loss. Therefore, computing top-$k$ token indices only at the final layer is sufficient, avoiding redundant per-layer selection and significantly improving efficiency in AR video generation. We incorporate our SEH into the final layer of AR video diffusion.

Under the attention mechanism~\cite{vaswani2017attention}, a token contributes meaningfully only when it is relevant, well-aligned, and informative, as encoded by the query, key, and value, respectively. Relying on only a subset of these components leads to incomplete salience estimation. Consequently, we concatenate QKV as the input to SEH to provide a joint and informative representation for salience prediction. Salience estimation can be formulated as a token-wise scalar regression task, where scores are used solely for ranking and cache retention. Moreover, QKV features are extracted after multiple Transformer layers and are already highly contextualized, obviating the need for additional spatial or temporal modeling. Accordingly, we adopt a lightweight MLP, termed Salience-MLP (S-MLP), as SEH, achieving a favorable trade-off between expressiveness and efficiency without introducing complex architectures such as attention or convolution. We further analyze the impact of the architectural choices through ablation studies.

\subsection{Training and Inference of PaFu-KV}
\label{sec:3.4}

Based on our analysis in Sec.~\ref{sec3.3}, we incorporate the S-MLP into the final layer of the AR video diffusion and activate it only at the $t_0$ timestep. At this stage, we concatenate the Q, K, and V representations along the feature dimension and pass the concatenated features to S-MLP to predict a scalar salience score for each token. With the estimated salience scores, PaFu-KV operates under two distinct regimes: training and inference, as indicated in Fig.~\ref{fig:framework}(b) and (c).

Our training procedure is built on the DMD pipeline. We calculate salience score $s_{\mathrm{real}}$ from real score network as supervision based on Eq.~\ref{eq4} and~\ref{eq5}. In parallel, we estimate salience scores from the generator $G_\theta$ using the SEH $S_{\phi}$. Our salience score loss is defined as the SmoothL1 loss between the predicted and the target salience scores, formulated as:
\begin{equation}
\mathcal{L}_{S}
= \mathrm{SmoothL1}\!\left(
S_{\phi}([Q,K,V]),\; s_{\mathrm{real}}
\right).
\end{equation}

During training, to reduce memory footprint, we enable the gradient checkpointing technique in PyTorch~\cite{paszke2019pytorch}, which avoids storing intermediate activations during the forward pass and recomputes them during backpropagation. In-place modification of the KV Cache leads to incorrect gradients due to inconsistent cache states across recomputed process. To address this issue, we adopt an index-based cache management strategy. Specifically, instead of explicitly reorganizing the KV Cache, we maintain a token index cache $I_s$, which records the indices of the top-$k$ salient tokens selected for each generated chunk based on both current and historical salience scores. The index cache is updated and rolled out synchronously with the KV Cache, ensuring consistent correspondence between token indices and KV entries throughout training. Consequently, Eq.~\ref{eq2} is modified as follows:
\begin{equation}
x_{t_{j-1}}^i
= \Psi\!\left(
G_\theta\!\left(x_{t_j}^i, t_j, \mathrm{Select}(KV; I_s)\right),
t_{j-1}
\right),
\end{equation}

\begin{figure*}[ht]
  \centering
  \includegraphics[width=\linewidth]{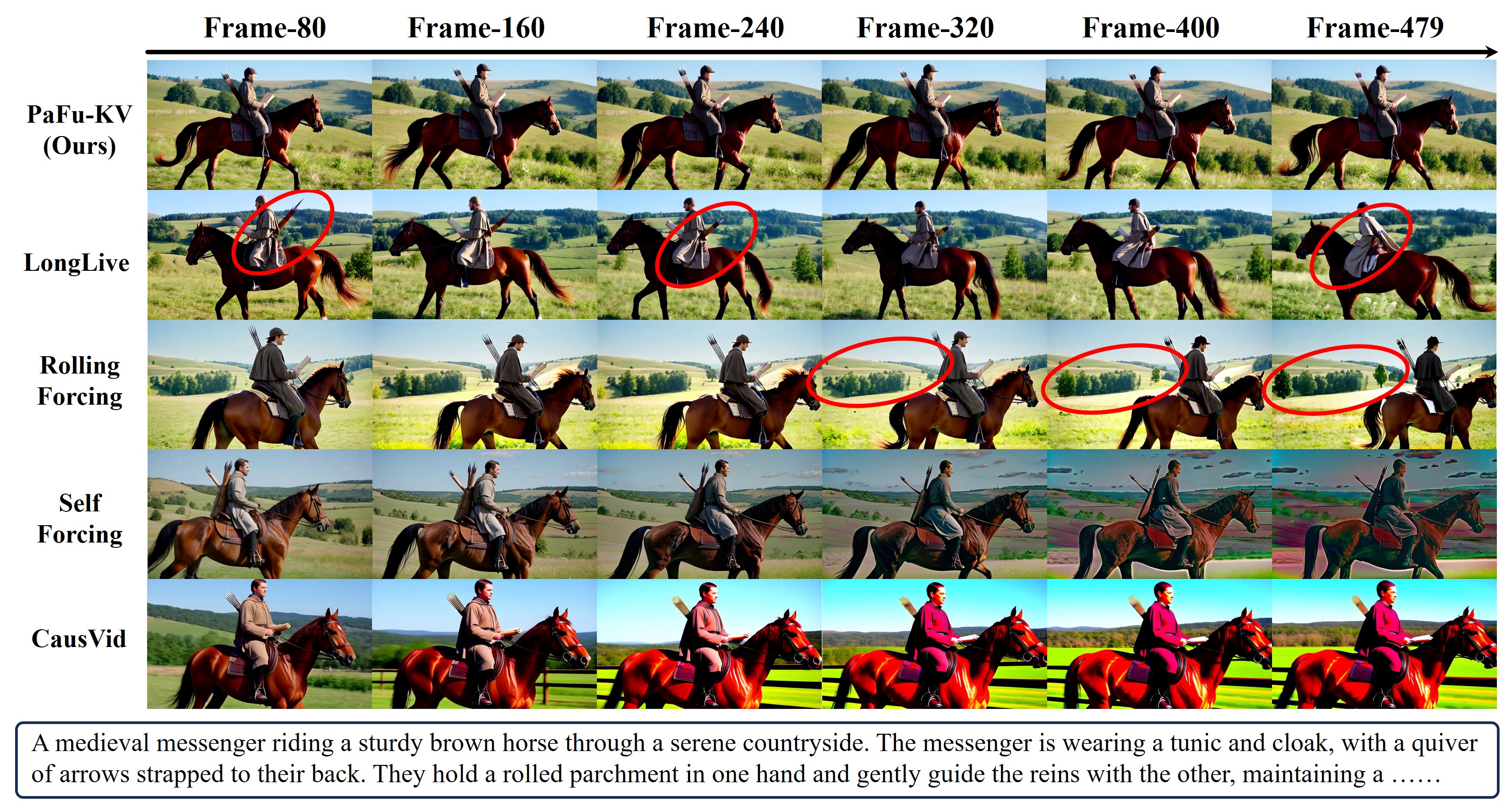}
  \caption{Qualitative experimental results on 30-second videos. We compare PaFu-KV with representative open-source autoregressive video generation models. We explicitly mark the inconsistent regions using a red circle in the figure for better visualization. }
\label{fig:cmp1}
\end{figure*}

During inference, we initialize an empty KV Cache and perform AR video generation in a chunk-wise fashion. After each chunk is generated, we obtain its corresponding KV entries and predicted salience scores. Once KV Cache reaches its capacity, we concatenate the salience scores of the current chunk with those of the historical cache and retain the top-$k$ salient tokens to update the KV Cache, evicting the remaining entries. The next chunk is generated conditioned on the updated cache, and this procedure is iteratively applied until the full video is generated.

\begin{table*}[ht]
\caption{Comparison results on short video generation. Experiments are conducted on a single H100 GPU, and evaluation scores are calculated on the official prompt of VBench~\cite{huang2024vbench}. Best results are highlighted in bold, second-best results are underlined.}
\centering
\small
\setlength{\tabcolsep}{3.2pt}
\renewcommand{\arraystretch}{1}
\begin{tabular}{llcccccc}
\toprule
\multicolumn{1}{c}{\multirow{1}{*}[-2ex]{\textbf{Model Types}}} & \multicolumn{1}{c}{\multirow{1}{*}[-2ex]{\textbf{Methods}}} &
\multicolumn{3}{c}{Real-time Performance} &
\multicolumn{3}{c}{Evaluation Scores} \\
\cmidrule(lr){3-5}\cmidrule(lr){6-8}
& & Params & Resolution & Throughput (FPS)$\uparrow$ & Total$\uparrow $ & Quality$\uparrow$ & Semantic$\uparrow$ \\
\midrule

\multirow{2}{*}{Bidirectional Diffusion} 
& LTX-Video\cite{hacohen2024ltx}          & 1.9B & $768 \times 512$ & 8.92 & 80.00 & 82.30 & 70.79 \\
& Wan\cite{wan2025wan}                    & 1.3B & $832 \times 480$ & 0.78  & 84.26 & 85.30 & 80.09 \\
\midrule
\multicolumn{1}{c}{\multirow{9}{*}{AR Diffusion}}
& SkyReels-V2\cite{chen2025skyreels}      & 1.3B & $960 \times 540$ & 0.49   & 82.67 & 84.70 & 74.53 \\
& MAGI-1\cite{teng2025magi}               & 4.5B & $832 \times 480$ & 0.19   & 79.18 & 82.04 & 67.74 \\
& CausVid\cite{yin2025slow}               & 1.3B & $832 \times 480$ & 17.0   & 81.20 & 84.05 & 69.80 \\
& NOVA\cite{deng2024autoregressive}       & 0.6B & $768 \times 480$ & 0.88   & 80.12 & 80.39 & \underline{79.05} \\
& Pyramid Flow\cite{jin2024pyramidal}     & 2.0B   & $640 \times 384$ & 6.7    & 81.72 & 84.74 & 69.62 \\
& Self Forcing\cite{huang2025self}        & 1.3B & $832 \times 480$ & 17.0   & \underline{84.31} & 85.07 & \textbf{81.28} \\
& LongLive\cite{yang2025longlive}         & 1.3B & $832 \times 480$ & 20.7   & 84.87 & \underline{86.97} & 76.47 \\
& Rolling Forcing\cite{liu2025rolling}    & 1.3B & $832 \times 480$ & 15.79  & 81.22 & 84.08 & 69.78 \\
& Ours                                    & 1.3B & $832 \times 480$ & \textbf{23.6} & \textbf{85.12} & \textbf{88.56} & 71.36 \\

\bottomrule
\end{tabular}
\label{tab:short}
\end{table*}

\begin{table*}[ht]
\caption{Comparison results on long video generation. Evaluation scores are calculated on VBench-Long~\cite{huang2025vbench++}. Best results are highlighted in bold, and second-best results are underlined.}
\centering
\small
\setlength{\tabcolsep}{7.5pt}
\renewcommand{\arraystretch}{1}
\begin{tabular}{lcccccc}
\toprule
\multicolumn{1}{c}{\multirow{1}{*}[-3ex]{\textbf{Methods}}} &
\multicolumn{6}{c}{Evaluation Scores} \\
\cmidrule(lr){2-7}
& \makecell{Subject\\consistency }$\uparrow$
& \makecell{Background\\consistency} $\uparrow$
& \makecell{Motion\\smoothness} $\uparrow$
& \makecell{Aesthetic\\quality} $\uparrow$
& \makecell{Imaging\\quality} $\uparrow$
& \makecell{Quality\\drift} $\uparrow$\\
\midrule
SkyReels-V2\cite{chen2025skyreels}      & 89.23 & 93.45 & 98.76 & 61.55 & 62.90 & 5.59 \\
MAGI-1\cite{teng2025magi}               & 90.86 & 93.25 & \textbf{99.20} & 59.91 & 59.87 & 2.15 \\
CausVid\cite{yin2025slow}               & 87.99 & 89.99 & 98.09 & 60.95 & 66.38 & 2.18 \\
Self Forcing\cite{huang2025self}        & 86.48 & 90.29 & 98.47 & 60.54 & 68.68 & 1.66 \\
LongLive\cite{yang2025longlive}         & 90.12 & 91.42 & \underline{99.07} & \textbf{62.94} & 69.46 & 0.59 \\
Rolling Forcing\cite{liu2025rolling}    & \underline{92.80} & \underline{93.71} & 98.70 & 62.39 & \underline{70.75} & \textbf{0.01} \\
Ours                                    & \textbf{93.91} & \textbf{94.14} & 98.14 & \underline{62.48} & \textbf{71.23} & \underline{0.29} \\
\bottomrule
\end{tabular}
\label{tab:long}
\end{table*}

\section{Experiments}
\subsection{Experimental Settings}

\textbf{Experimental Implementation.} Our PaFu-KV was trained under the Self Forcing~\cite{huang2025self} paradigm. To enable long video generation, we implemented the streaming long tuning strategy proposed in Longlive~\cite{yang2025longlive}. Our PaFu-KV is built on Wan2.1-T2V-1.3B~\cite{wan2025wan}, a flow matching model based on the DiT architecture originally designed for generating 5-second clips, and distilled from Wan2.1-T2V-14B in DMD training pipeline. We train our model in two stages. In the first stage, we trained the SEH and AR video generator for 500 iterations without KV Cache eviction. In the second stage, we train both components for an additional 2,500 iterations with KV Cache eviction activated. Training runs for 3,000 iterations on 16 H200 GPUs with gradient accumulation of 4, spanning approximately two days. We optimize the AR generator and SEH using separate AdamW optimizers with the same learning rate of $1\times10^{-5}$. The critical models in the DMD pipeline are optimized using additional AdamW optimizers with a learning rate of $2\times10^{-6}$. All optimizers use the same momentum parameters, with $\beta_1=0.0$ and $\beta_2=0.999$. Our model is trained on extended VidProM prompts proposed in Longlive~\cite{yang2025longlive}. The KV Cache size is set to 4680, and ablation studies on the cache size are provided in Sec.~\ref{sec:abla}. Additionally, we incorporate the frame-sink mechanism from LongLive~\cite{yang2025longlive} and evaluate its effect via ablation. Additional training details are provided in the Supplementary Material.

\textbf{Evaluation Protocol.} We compare our method against a range of existing open-source video generation models under two evaluation protocols: short video generation and long video generation. For short video generation, we include LTXVideo~\cite{hacohen2024ltx}, Wan2.1~\cite{wan2025wan}, SkyReels-V2~\cite{chen2025skyreels}, MAGI-1~\cite{teng2025magi}, CausVid~\cite{yin2025slow}, NOVA~\cite{deng2024autoregressive}, Pyramid Flow~\cite{jin2024pyramidal}, Self Forcing~\cite{huang2025self}\, LongLive~\cite{yang2025longlive}, and Rolling Forcing. For long video generation, we compare with SkyReels-V2~\cite{chen2025skyreels}, CausVid~\cite{yin2025slow}, Self Forcing~\cite{huang2025self}, and LongLive~\cite{yang2025longlive}. For clarity and fairness, we categorize all baselines into two groups according to their generation paradigms: bidirectional diffusion models and AR models. Specifically, the bidirectional diffusion models include LTXVideo and Wan2.1-1.3B, while the remaining fall under the AR model category.

\textbf{Evaluation Metrics.} For short video generation, we generate 5-second videos using extended VBench~\cite{huang2024vbench} prompts following Self Forcing~\cite{huang2025self} and evaluate the results with VBench metrics, where the total score is computed as a weighted average of a quality score and a semantic score. For long video generation, we generate 30-second videos at 16 fps with a resolution of $832 \times 480$ using the official prompts from VBench-Long~\cite{huang2025vbench++}. We evaluate the generated videos using subject consistency, background consistency, motion smoothness, aesthetic quality, and imaging quality. Additionally, we compute the absolute difference in imaging quality to assess quality drift as proposed by~\citet{zhang2025packing}, and evaluate real-time performance in terms of generation throughput.

\subsection{Experimental Results}

\textbf{Quantitative Results.} Comprehensive evaluations are conducted on short and long video generation benchmarks, namely VBench~\cite{huang2024vbench} and VBench-Long~\cite{huang2025vbench++}. As shown in Table~\ref{tab:short}, for short video generation, our PaFu-KV achieves a higher total score than Self Forcing and LongLive, as well as a higher semantic score compared to Rolling Forcing. This improvement can be attributed to our selection of more salient tokens during generation. In addition, by reducing the KV Cache size, PaFu-KV outperforms existing methods in real-time performance, achieving higher throughput. Furthermore, as presented in Table~\ref{tab:long}, PaFu-KV achieves the best performance in subject and background consistency, while exhibiting lower quality drift with only marginal degradation in motion smoothness. Overall, quantitative results on both short and long video generation indicate that PaFu-KV effectively accelerates AR diffusion-based generation while improving temporal consistency with negligible quality degradation. 

\textbf{Qualitative Results.}
Qualitative analysis comparing our method with existing open-source approaches demonstrates the effectiveness of PaFu-KV. In particular, regions with noticeable inconsistencies are highlighted for clearer comparison. As shown in Fig.~\ref{fig:cmp1}, CausVid~\cite{yin2025slow} and Self Forcing~\cite{huang2025self} exhibit severe temporal drift caused by error accumulation when generating long videos. Rolling Forcing~\cite{liu2025rolling} suffers from background inconsistency across frames. In contrast, LongLive~\cite{yang2025longlive} fails to preserve subject consistency over long temporal horizons. Compared with these methods, our approach achieves substantially improved temporal consistency in long video generation and demonstrates comparable robustness to error accumulation, particularly when contrasted with CausVid and Self Forcing. Additional qualitative results are provided in the Supplementary Material. For qualitative results in video format, additional materials are available at: \url{https://anonymous.4open.science/r/PaFu-KV-B7EE/README.md}.

\subsection{Ablation Study}
\label{sec:abla}

To assess the contribution of key components in PaFu-KV, we conduct ablation experiments focusing on four critical factors: (1) the architecture of SEH, (2) the computation of the salience score, (3) the size of the reduced KV Cache, and (4) the implementation of frame sink. All ablation experiments are conducted on VBench-Long~\cite{huang2025vbench++} using the same evaluation protocol as long video generation. As summarized in Table~\ref{tab:abla}, both the S-MLP design in SEH and the proposed salience score computation have a significant impact on long video generation performance. For clarity, \textit{Subj.}, \textit{Back.}, \textit{Mot.}, \textit{Aes.}, \textit{Imag.}, and \textit{Drift} denote subject consistency, background consistency, motion smoothness, aesthetic quality, imaging quality, and quality drift, respectively. Visualization results of the ablation study are provided in the Supplementary Material.

\textbf{The impact of SEH architecture.} We evaluate the impact of the SEH architecture by implementing SEH with a multi-head attention (MHA) mechanism and a 1D convolution configured with a kernel size of 3. As indicated in Table~\ref{tab:abla}, despite incorporating a larger number of parameters, the generation quality degrades, suggesting that increased architectural complexity does not improve long video generation.

\textbf{The impact of salience score computation.} To examine the effectiveness of our proposed salience score computation, we replace it with simpler alternatives, including global maximum pooling and global average pooling.‌ As indicated in Table~\ref{tab:abla}, global maximum pooling yields poorer subject and background consistency and higher quality drift, while global average results in severe degradation. 

\textbf{The impact of reduced KV Cache size.} We further investigate the effect of reducing the KV Cache size, where the cache size of the full model is 4680. We evaluate our PaFu-KV under four cache size configurations: 9360, 6240, 3120, and 1560 tokens. Each configuration is a multiple of 1560, which corresponds to the number of latent tokens used to represent a single video frame. Experimental results show that cache sizes of 9360 and 6240 tokens achieve performance comparable to the full model. However, when the cache size is reduced below 4680 tokens, a noticeable degradation in generation quality is observed. 

\textbf{The impact of frame sink.} We conduct an ablation study on the frame-sink mechanism. The results indicate that removing frame sink leads to marginal degradation in long video generation performance, suggesting that it is not a primary contributor to the performance gains of PaFu-KV.

\begin{table}[ht]
\caption{Ablation experiments are conducted on VBench-Long following the long video generation protocol.}
\centering
\small
\setlength{\tabcolsep}{0.9pt}
\renewcommand{\arraystretch}{1.2}
\begin{tabular}{llcccccc}
\toprule
\multicolumn{1}{c}{\textbf{Factors}} & \multicolumn{1}{c}{\textbf{Settings}} &
\multicolumn{6}{c}{Evaluation scores} \\
\cmidrule(lr){3-8}
& 
& \makecell{Subj.$\uparrow$} 
& \makecell{Back.$\uparrow$} 
& \makecell{Mot.$\uparrow$} 
& \makecell{Aes.$\uparrow$} 
& \makecell{Imag.$\uparrow$} 
& \makecell{Drift$\uparrow$} \\
\midrule
\multicolumn{1}{c}{\multirow{2}{*}{\makecell{SEH \\ architecture}}}
  & \multicolumn{1}{c}{Conv} & 85.47 & 85.13 & 94.38 & 58.04 & 67.04 & 1.68 \\
  & \multicolumn{1}{c}{MHA} & 88.65 & 88.71 & 95.74 & 60.24 & 69.47 & 1.18 \\
\midrule
\multicolumn{1}{c}{\multirow{2}{*}{\makecell{Salience score \\ computation}}}
  & \multicolumn{1}{c}{Max} & 87.65 & 88.71 & 94.34 & 60.18 & 69.33 & 1.45 \\
  & \multicolumn{1}{c}{Avg} & 78.87 & 79.78 & 76.94 & 52.94 & 58.49 & 2.54 \\
\midrule
\multicolumn{1}{c}{\multirow{4}{*}{\makecell{KV Cache\\size}}}
  & \multicolumn{1}{c}{9360} & 94.12 & 93.89 & 98.04 & 62.06 & 71.89 & 0.25 \\
  & \multicolumn{1}{c}{6240} & 93.45 & 94.29 & 98.54 & 62.57 & 70.76 & 0.31 \\
  & \multicolumn{1}{c}{3120} & 85.26 & 86.48 & 82.11 & 53.48 & 64.58 & 1.12  \\
  & \multicolumn{1}{c}{1560} & 74.48 & 75.18 & 79.44 & 49.74 & 58.69 & 2.36 \\
\midrule
\multicolumn{1}{c}{\multirow{1}{*}[0.9ex]{\makecell{Frame\\sink}}}
  & \multicolumn{1}{c}{\makecell{w/o\\frame sink}}  & 93.12 & 93.09 & 97.84 & 61.89 & 70.45 & 0.44 \\
\midrule
\multicolumn{1}{c}{Full Model} & \multicolumn{1}{c}{-} & \textbf{93.91} & \textbf{94.14} & \textbf{98.14} & \textbf{62.48} & \textbf{71.23} & \textbf{0.29 } \\

\bottomrule
\end{tabular}
\label{tab:abla}
\end{table}

\section{Conclusion}

We introduce PaFu-KV, a past- and future-informed KV Cache policy guided by a principled salience estimation framework. By analyzing attention patterns in bidirectional diffusion models, we identify token salience as a key factor governing long-horizon generation stability. Our proposed salience score explicitly disentangles attention contributions from past, present, and future interactions, effectively mitigating the strong diagonal bias inherent in self-attention and enabling a more balanced assessment of token importance across temporal spans. We then design a lightweight Salience Estimation Head (SEH) that distills global salience information from a bidirectional teacher model into an AR video diffusion model under causal constraints. During inference, PaFu-KV leverages the predicted salience scores to selectively retain informative tokens while evicting those with marginal generative utility, thereby maintaining a compact KV Cache and reducing interference from invalid or drifting representations. Extensive experiments on both short- and long-horizon video generation benchmarks demonstrate that PaFu-KV substantially reduces KV Cache size and accelerates inference while preserving high-fidelity video generation. Moreover, our method consistently improves temporal coherence and achieves comparable error accumulation in long video synthesis.

\nocite{langley00}

\bibliography{example_paper}
\bibliographystyle{icml2026}

\newpage
\appendix
\onecolumn

\section{Additional Implementation Details}

\textbf{KV Cache and Salience Score List.} In the full PaFu-KV model, we set the KV Cache size to 4680 tokens, corresponding to three times the number of latent tokens per video frame, and use this configuration consistently during training and inference. During training, we enable gradient checkpointing in PyTorch to reduce GPU memory footprint by avoiding the storage of intermediate activations and recomputing them during backpropagation. Each training iteration operates on a 5-second video segment, corresponding to 21 latent frames. As a result, the effective KV Cache size during training is required to be substantially larger than that used at inference. Specifically, the cache size is first expanded to $21 \times 1560 = 32{,}760$ tokens, corresponding to 21 latent frames. When adopting long video tuning following LongLive~\cite{yang2025longlive}, the KV Cache is further enlarged to $33 \times 1560 = 51{,}480$ tokens to support extended AR rollouts. We set the salience score list to have the same size as the KV Cache, storing 51,480 scalar values, while the KV Cache stores 51,480 tokens each with 12 attention heads and a feature dimension of 128. Consequently, the memory overhead of the salience score list is negligible compared to that of the KV Cache. As described in Sec.~\ref{sec:3.4}, we maintain an index structure to select the top-$k$ salient tokens for each denoising chunk. When a latent segment of 21 frames is denoised, both the DMD loss and the salience score loss are computed to optimize the AR video diffusion model. Although the KV Cache is configured with a large capacity during training, the effective KV Cache used for attention computation within each denoising chunk remains fixed at 4680 tokens, consistent with the inference-time setting. Since each timestep denoises a video chunk consisting of 4680 tokens, we set $L_B=4680$ as the block length in the Spatialtemporal-balanced Salience Estimation used by the real score network. During inference, the KV Cache size is accordingly set to 4680 tokens.

\textbf{S-MLP.} We adopt a lightweight MLP (S-MLP) as the architecture for the Salience Estimation Head (SEH). The S-MLP consists of two linear layers with a SiLU activation in between, and is incorporated into the final layer of the AR video diffusion model. The query, key, and value tensors, each of shape $\mathbb{R}^{B \times N \times L \times D}$ with $B = 1$, $N = 12$, $L = 4680$, and $D = 128$, are first concatenated. We then permute the tensor to exchange the $N$ and $L$ dimensions and reshape it by merging the attention-head dimension $N$ with the feature dimension $D$, resulting in a merged feature dimension $D_M = N \times D = 1536$. The resulting query, key, and value representations are further concatenated along the $D_M$ dimension and fed into the S-MLP. The input, hidden, and output dimensions of the S-MLP are 4608, 1024, and 12, respectively. Finally, an averaging operation is applied over the output dimension of size 12 to produce the salience score list, which is subsequently used for salience-based token selection and KV Cache management.

\section{Streaming Long Tuning and  KV Re-caching in LongLive.} 
\textbf{Streaming Long Tuning.} Our PaFu-KV is trained using Streaming Long Tuning proposed by~\citet{yang2025longlive}. Streaming Long Tuning is designed to address the error accumulation that arises when AR video models trained on short clips are deployed for long-horizon generation. Instead of adopting a train-short–test-long regime, this procedure explicitly aligns training with inference by simulating long AR video diffusion rollouts. Concretely, the training proceeds in a streaming manner: the model first generates a short video clip from an empty context and receives supervision from a teacher model (Wan2.1-14B) on the generated clip. The generated frames are then stored in the KV Cache and treated as fixed historical context for the next iteration, in which the model generates the subsequent clip conditioned on its own past predictions. This rolling process continues until a predefined maximum video length is reached, after which training restarts from a fresh context. To ensure scalability, gradients are computed only for the newly generated clip at each iteration, while earlier frames are detached and used solely as causal context, thereby bounding memory usage by the clip length rather than the total video duration. By repeatedly exposing the model to long, self-generated, and progressively imperfect contexts during training, streaming long tuning mitigates error accumulation and temporal drift, and effectively aligns the training distribution with inference-time conditions.

\textbf{KV Re-caching.} KV re-caching is introduced to facilitate smooth and semantically consistent prompt switching during interactive long video generation. In causal AR models, prompt information is repeatedly injected through cross-attention layers and propagated forward by self-attention, causing prompt semantics to be deeply embedded in the KV Cache. As a result, simply retaining the cache across prompt switches often leads to delayed or incorrect adherence to new prompts, whereas clearing the cache entirely introduces abrupt visual discontinuities. KV re-caching resolves this trade-off by selectively refreshing the cache at prompt boundaries. Specifically, at the first generation step following a prompt switch, the model recomputes the KV Cache by jointly encoding the already generated video frames as visual context together with the new prompt, thereby removing residual semantics from the previous prompt while preserving motion and appearance continuity. Subsequent frames are then generated autoregressively using this refreshed cache without further intervention. This re-caching operation is performed only once per prompt switch and incurs minimal additional overhead. By integrating KV re-caching into both training and inference, the model learns to rapidly adapt to new prompts while maintaining smooth temporal transitions, ensuring consistent visual quality and prompt adherence in interactive long video generation.

\section{Algorithm}

To facilitate reproducibility and provide a precise procedural description of our method, we present the complete algorithmic workflow of the training and inference of PaFu-KV in Alg.~\ref{alg:train} and~\ref{alg:inference}, respectively. Additionally, computing the ground-truth salience scores in the teacher model requires explicit calculation of attention weights, which can incur substantial GPU memory overhead during training, as the teacher employs full bidirectional self-attention and involves all tokens in the attention computation. To mitigate this issue, we adopt a block-wise attention computation strategy to reduce memory footprint, as detailed in Algorithm~\ref{alg:salience}. The pseudocode summarizes the training and inference procedures described in Sec.~\ref{sec:3.4}, including KV Cache management, salience score computation, and token selection. All operations adhere to the notations and settings defined in the main text.

\begin{figure*}[t]
  \centering
  \includegraphics[width=\linewidth]{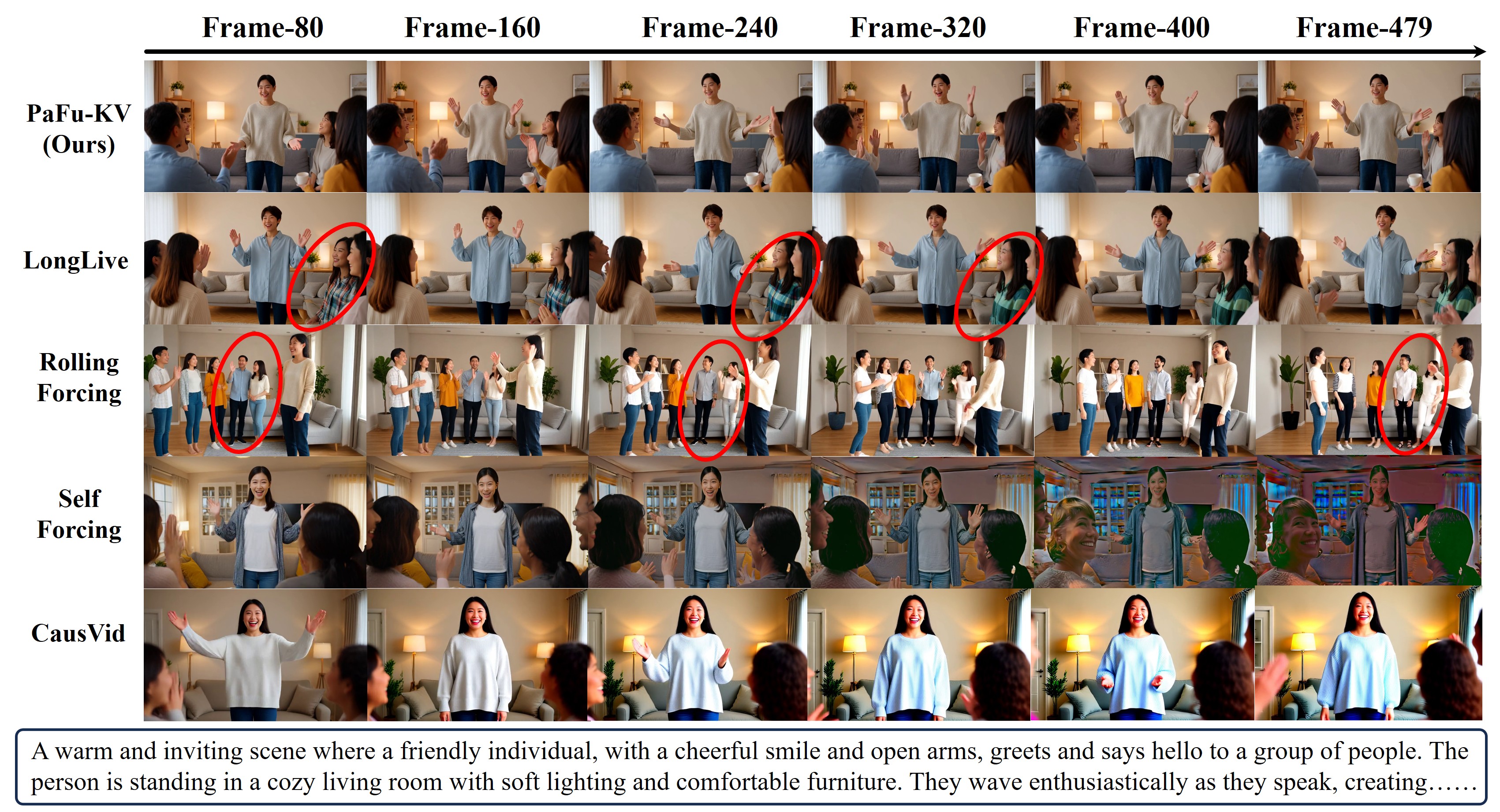}
  \caption{Qualitative experimental results on 30-second videos. We compare PaFu-KV with representative open-source autoregressive video generation models. We explicitly mark the inconsistent regions using a red circle in the figure for better visualization. }
\label{fig:cmp2}
\end{figure*}

\begin{figure*}[t]
  \centering
  \includegraphics[width=\linewidth]{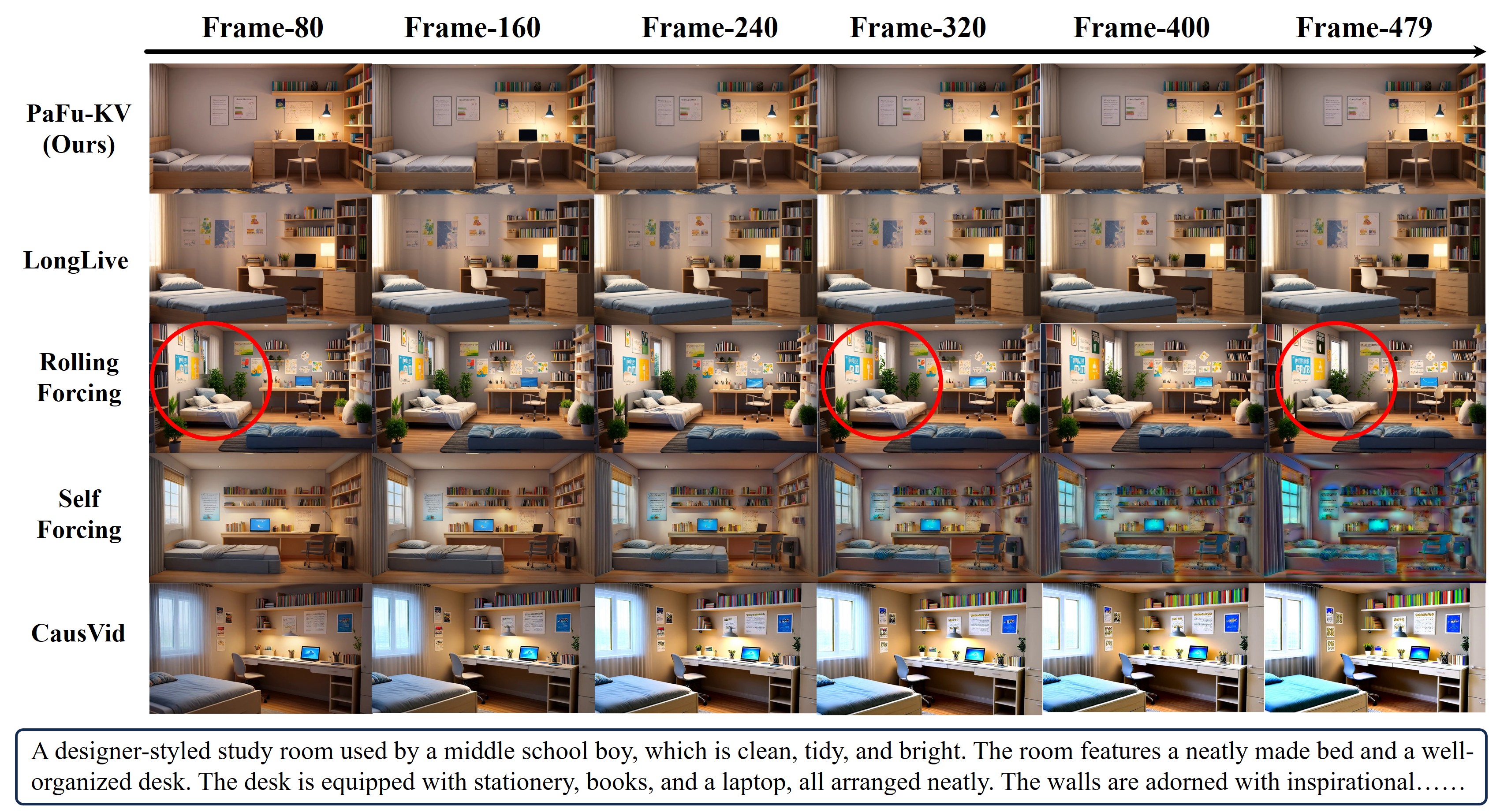}
  \caption{Qualitative experimental results on 30-second videos. We compare PaFu-KV with representative open-source autoregressive video generation models. We explicitly mark the inconsistent regions using a red circle in the figure for better visualization. }
\label{fig:cmp3}
\end{figure*}

\begin{figure}
  \centering
  \includegraphics[width=\linewidth]{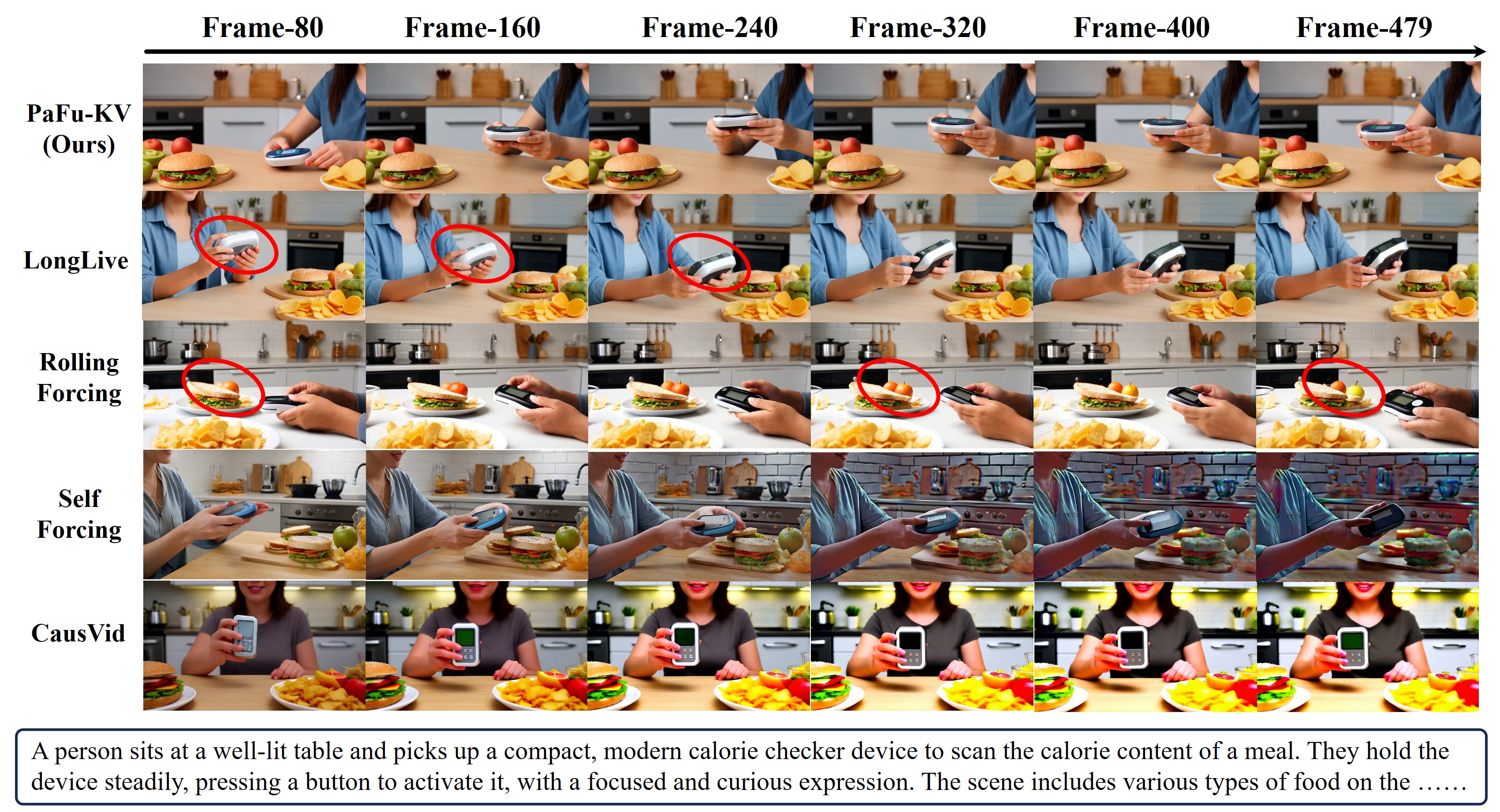}
  \caption{Qualitative experimental results on 30-second videos. We compare PaFu-KV with representative open-source autoregressive video generation models. We explicitly mark the inconsistent regions using a red circle in the figure for better visualization. }
\label{fig:cmp4}
\end{figure}

\begin{figure}
  \centering
  \includegraphics[width=\linewidth]{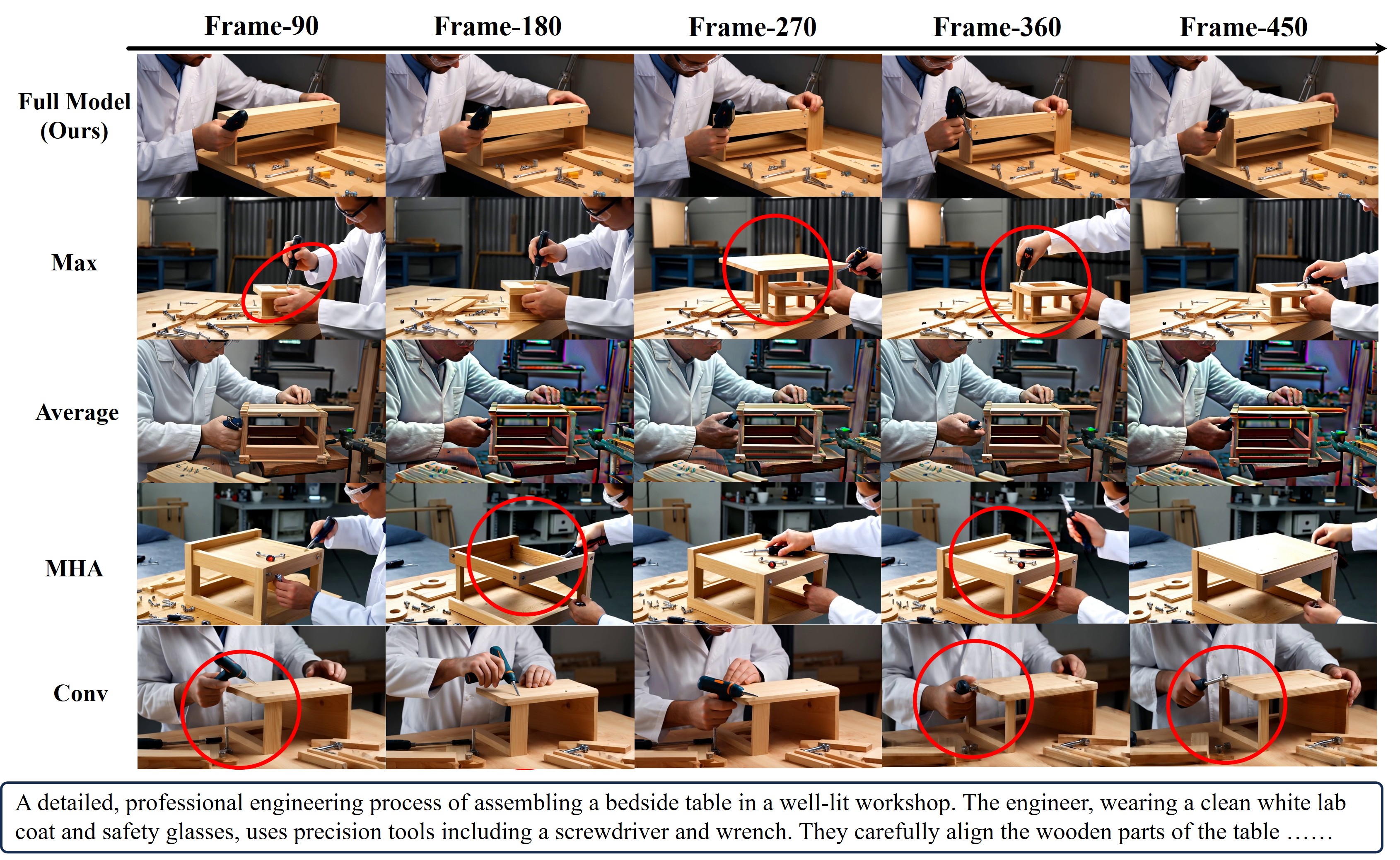}
  \caption{Visualization of ablation study. }
\label{fig:abla}
\end{figure}

\section{Additional Qualitative Results.}

In this section, we provide additional qualitative visual comparisons in Fig.~\ref{fig:cmp2}, Fig.~\ref{fig:cmp3}, and Fig.~\ref{fig:cmp4}. The prompt used for each example is shown below the corresponding figure, and regions exhibiting noticeable inconsistencies are highlighted using red circles. The results indicate that our method achieves strong long-horizon consistency. Qualitative results in video format are provided at \url{https://anonymous.4open.science/r/PaFu-KV-B7EE/README.md}. Additionally, we provide visualizations for the ablation study in Fig~\ref{fig:abla}. As illustrated in the figure, the averaging strategy results in substantial error accumulation, whereas attention-based and naive max strategies markedly reduce temporal consistency.

\section{Discussion and Future Work.}
In this section, we provide further discussion of PaFu-KV and outline directions for future work. From a practical deployment perspective, in addition to generation speed, GPU memory footprint is a critical factor when deploying video generation models. Although PaFu-KV reduces the KV Cache size by selecting top-$k$ salient tokens and thereby accelerates inference, the resulting reduction in memory usage is relatively limited. This phenomenon is common among KV-cache optimization approaches in the AR video diffusion domain. The primary reason is the widespread adoption of FlashAttention~\cite{dao2022flashattention}, which significantly reduces the memory footprint of the attention mechanism, the dominant source of memory consumption in Transformer architectures. Specifically, FlashAttention computes attention in a block-wise manner using a streaming softmax, discarding intermediate $QK^\top$ scores and softmax probabilities on the fly instead of materializing the full attention matrix. As a result, it trades recomputation for memory efficiency and reduces the memory complexity from $O(N^2)$ to approximately $O(N)$. Under this regime, further reducing the KV Cache size yields only marginal memory savings. Consequently, reducing GPU memory footprint most effectively requires reducing the model parameter. In future work, we plan to further investigate AR video diffusion models with the goal of improving deployment efficiency while maintaining high generation quality, thereby enhancing the practical applicability of long video generation systems.

\twocolumn

\begin{algorithm}[tb]
\caption{PaFu-KV Training}
\label{alg:train}
\begin{algorithmic}[1]
\Require Denoise timesteps $\{t_0,t_1,\ldots,t_T\}$
\Require Number of video frames $N$
\Require AR diffusion model $G_\theta$ (returns KV embeddings via $G_\theta^{\mathrm{KV}}$)
\Require Salience MLP $S_{\hat{\theta}}$ (takes intermediate features and outputs salience score)
\Require Token number of top salience score $k$

\Statex \textbf{loop}
\Statex Initialize KV Cache $\mathbf{KV} \gets [\,]$
\Statex Initialize generator salience score list $\mathbf{SL_g} \leftarrow [\,]$
\Statex Initialize list of indices for selected top-$k$ token in KV Cache $\mathbf{IL} \leftarrow [\,]$
\Statex Sample $s \sim \mathrm{Uniform}(0,2,\ldots,T)$

\For{$i = 1, \ldots, N$}
    \State Initialize $x^{i}_{t_T} \sim \mathcal{N}(0, I)$
    \For{$j = T, \ldots, s$}
        \If{$j = s$}
            \State Enable gradient computation
            \State Set $\hat{x}^{i}_0 \gets G_\theta(x^{i}_{t_j};\, t_j,\, \text{Select}(\mathbf{KV};\mathbf{IL}))$
            \State Enable salience computation
            \State Set$\{\!q,\!k,\!v\!\},\mathbf{kv}^i\!\!\gets\!\!G_\theta(\hat{x}^{i}_0;\!0,\!\text{Select}(\!\mathbf{KV};\!\mathbf{IL}\!)\!)$
            \State Set $s^{i}_g \gets S_{\hat{\theta}}(\{q,k,v\})$
            \State $\mathbf{KV}.\mathrm{append}(\mathbf{kv}^i)$
            \State $\mathbf{SL_g}.\mathrm{append}(s^i_g)$
            \State Indices $ids^i \gets \text{Top-$k$}(\mathbf{SL_g}) $ 
            \State $\mathbf{IL}.\mathrm{append}(ids^i)$
            \State Disable salience computation
            \State Disable gradient computation
        \Else
            \State Disable gradient computation
            \State Set $\hat{x}^{i}_0 \gets G_\theta(x^{i}_{t_j};\, t_j,\, \text{Select}(\mathbf{KV};\mathbf{IL}))$
            \State Sample $\epsilon \sim \mathcal{N}(0, I)$
            \State Set $x^{i}_{t_{j-1}} \gets \Psi(\hat{x}^{i}_0, \epsilon, t_{j-1})$
        \EndIf
    \EndFor
\EndFor
\Statex Compute distribution matching loss $\mathcal{L}_{\mathrm{dm}}$
\Statex Compute salience score list in real score network $\mathbf{SL}_r$
\Statex Set $\mathcal{L}_{\mathrm{s}} \gets \mathrm{SmoothL1}\!\left(\mathbf{SL}_r,\mathbf{SL_g}\right)$
\Statex Update $\theta$ by minimizing $\mathcal{L}_{\mathrm{dm}}$ and $\mathcal{L}_{\mathrm{s}}$
\Statex \textbf{end loop}
\end{algorithmic}
\end{algorithm}

\begin{algorithm}[tb]
\caption{PaFu-KV Inference}
\label{alg:inference}
\begin{algorithmic}[1]
\Require KV Cache size of $L$ tokens
\Require Denoise timesteps $\{t_0,\ldots,t_T\}$
\Require Number of generated video frames $M$
\Require AR diffusion model $G_\theta$, salience MLP $S_{\hat{\theta}}$
\Require Token number of top salience score $k$

\Statex Initialize KV Cache $\mathbf{KV} \gets [\,]$
\Statex Initialize salience score list $\mathbf{SL} \leftarrow [\,]$
\Statex Sample $s \sim \mathrm{Uniform}(0,1,\ldots,T)$

\For{$i = 1, \ldots, M$}
    \State Initialize $x^{i}_{t_T} \sim \mathcal{N}(0, I)$
    \For{$j = T, \ldots, 0$}
        \State Set $\hat{x}^{i}_0 \gets G_\theta(x^{i}_{t_j};\, t_j,\, \text{Select}(\mathbf{KV};\mathbf{IL}))$
        \If{$j = 0$}
            \State Set $\mathbf{kv}^i, s^i\!\!\gets\!\!G_\theta(\hat{x}^{i}_0;\!0,\!\text{Select}(\!\mathbf{KV};\!\mathbf{IL}\!)\!)$
            \If{$\|\mathbf{KV}\| = L$}
                \State Indices $ids^i \gets \text{Top-$k$}(\mathbf{SL_g}) $ 
                \State $\mathbf{KV} \gets \text{Select}(\mathbf{KV};\, ids^i)$
                \State $\mathbf{SL} \gets \text{Select}(\mathbf{SL};\, ids^i)$
            \EndIf
            \State $\mathbf{KV}.\mathrm{append}(\mathbf{kv}^i)$
            \State $\mathbf{SL}.\mathrm{append}(\mathbf{s}^i)$
        \Else
            \State Set $x^{i}_{t_{j-1}}$ via $G_\theta$ with $\text{Select}(\mathbf{KV};\mathbf{IL})$
        \EndIf
    \EndFor
\EndFor
\end{algorithmic}
\end{algorithm}

\begin{algorithm}[t]
\caption{Block-wise Salience Score Computation}
\label{alg:salience}
\begin{algorithmic}[1]
\Require Query, Key $\mathbf{Q}$, $\mathbf{K}$
\Require Chunk size $c$, block size $b$
\Ensure Salience score $\mathbf{s}$
\State Initialize $\mathbf{out}_{\text{upper}}, \mathbf{out}_{\text{diag}}, \mathbf{out}_{\text{down}} \gets \mathbf{0}$
\For{$s = 0$ \textbf{to} $N-1$ \textbf{step} $c$}
    \State $e \gets s + c$
    \State $i \gets \lfloor s / b \rfloor$ \Comment{Block index}
    \State $l \gets i \cdot b$, \quad $h \gets l + b$
    \State Extract query chunk $\mathbf{Q}_{s:e}$
    \State Compute attention weights $\mathbf{A} \gets \mathbf{Q}_{s:e} \mathbf{K}^\top /\sqrt{D}$
    \State Aggregate salience $\mathbf{m} \gets \max \; \mathbb{E}_{\text{head}}[\mathbf{A}]$
    \State $\mathbf{s}_{\text{down}}[:, \!:\!l] \gets \max(\mathbf{out}_{\text{down}}[:, :\!l], \mathbf{m}[:, :\!l])$
    \State $\mathbf{s}_{\text{diag}}[:, \!l\!:\!h] \gets \max(\mathbf{out}_{\text{diag}}[:, l\!:\!h], \mathbf{m}[:, l\!:\!h])$
    \State $\mathbf{s}_{\text{upper}}[:, \!h\!:] \gets \max(\mathbf{out}_{\text{upper}}[:, h\!:], \mathbf{m}[:, h\!:])$
\EndFor

\State Initialize $\mathbf{out} \in \mathbb{R}^{B \times N}$
\State Fuse salience scores:
\State \Return $\mathbf{out}$
\end{algorithmic}
\end{algorithm}

\onecolumn

\end{document}